\pdfoutput=1

\documentclass[11pt]{article}
\usepackage[most]{tcolorbox}

\usepackage[preprint]{acl}

\usepackage{times}
\usepackage{latexsym}

\usepackage[T1]{fontenc}

\usepackage[utf8]{inputenc}

\usepackage{microtype}

\usepackage{inconsolata}

\usepackage{setspace}  
\usepackage{times}  

\usepackage{hyperref}
\usepackage{rotating}  
\usepackage{linguex}
\usepackage{tcolorbox}
\usepackage{lmodern}

\usepackage{graphicx}

\usepackage[table,xcdraw]{xcolor}

\usepackage{lscape}
\usepackage{stmaryrd}
\usepackage{tabto}
\usepackage{tikz-qtree,tikz-qtree-compat}
\usepackage{linguex}

\usepackage{ragged2e} 
\usepackage{mathptmx}  
\usepackage{amsmath}  
\usepackage{ccg}  
\usepackage{amsmath}
\usepackage{ccg-latex}
\usepackage{tablefootnote}
\newcommand{\sembrac}[1]{$\llbracket\textrm{#1}\rrbracket$}
\newcommand{\tset}[1]{$\langle\textrm{\textit{#1}}\rangle$}

\usepackage{multirow}

\usepackage{geometry}
\usepackage{mathtools, nccmath}
\usepackage{enumitem}

\usepackage{multirow}
\usepackage{graphicx}
\usepackage{caption}

\title{FSLI: An Interpretable Formal Semantic System for\\
 One-Dimensional Ordering Inference}

\author{
  Maha Alkhairy, Vincent Homer, Brendan O'Connor \\
  University of Massachusetts Amherst, Amherst, MA, USA \\
  \texttt{malkhairy@umass.edu} 
}

\begin{document}
\maketitle
\begin{abstract}

We develop a system for solving logical deduction one-dimensional ordering problems by transforming natural language premises and candidate statements into first-order logic. Building on \citet{Heim1998SemanticsIG}'s syntax-based compositional semantic rules which utilizes lambda calculus, we develop a semantic parsing algorithm with abstract types, templated rules, and a dynamic component for interpreting entities within a context constructed from the input. The resulting logical forms are executed via constraint logic programming to determine which candidate statements can be logically deduced from the premises.

The symbolic system, the Formal Semantic Logic Inferer (FSLI), provides a formally grounded, linguistically driven system for natural language logical deduction. We evaluate it on both synthetic and derived logical deduction problems. FSLI achieves 100\% accuracy on BIG-bench’s \textit{logical\_deduction} task and 88\% on a syntactically simplified subset of AR-LSAT outperforming an LLM baseline, o1-preview. 

While current research in natural language reasoning emphasizes neural language models, FSLI highlights the potential of  principled, interpretable systems for symbolic logical deduction in NLP. 

\end{abstract}

\section{Introduction}

Natural language processing (NLP) methods have made impressive recent advances in various tasks. However, they continue to struggle with deductive inference and logical deduction, which often requires the incorporation of computational logic.

We investigate an important class of deductive logical problems about one-dimensional ordering, where a natural language paragraph describes aspects of how a set of discrete items are ordered (e.g., left to right, or through time).  The `logical\_deduction' task from the BIG-bench LLM evaluation suite (\citet{srivastava2023beyond}, Section \ref{sec:big_bench}) exclusively consists of such ordering problems, constructed in a multiple choice format, such as Example \ref{ex:bigbench}:

\begin{tcolorbox}[
    colback=white!10,
    colframe=gray!75!black,
    sharp corners=all,
    boxrule=0.1mm,
    left=-0.1mm
]

\small
\refstepcounter{example}\label{ex:bigbench}
\noindent\textbf{Example~{\theexample} (from BIG-bench logical\_deduction):}\\ 

\textbf{Premise:} On a branch, there are three birds: a raven, a quail, and a crow. The quail is the leftmost. The raven is the rightmost. \\

\textbf{Candidate Statements:}
{[}`The raven is the second from the left.', `The quail is the second from the left.', `The crow is the second from the left.'{]}
\end{tcolorbox}

More difficult one-dimensional ordering problems are also available from 
from the `logic games' section of the Law School Admission Test (LSAT), used to evaluate human abilities at logical deduction. These tasks are linguistically and logically demanding, requiring models to reason over constraints to determine valid orderings over a combinatorially large space, often with multiple correct solutions (Section~\ref{sec:lsat}). 

Existing approaches to logical deduction in NLP include large  language models (LLMs) \cite{openai2023gpt4, touvron2023llama, anil2023palm2}, hybrid systems that integrate LLMs with logic engines \cite{kirtania-etal-2024-logic, PanLogicLM23}, and semantic parsers based on Combinatory Categorial Grammar (CCG) \cite{Steedman2019Combinatory}. 

While each of these approaches are viable, they also have limitations (detailed in Section \ref{sec:relevantlit}); for example, LLMs are pattern-based and prone to logical inconsistencies.  
We investigate an alternative approach for solving logical deduction problems, to adopt symbolic systems that incorporate dedicated logic inference modules \cite{cooper1993prolog, pereira2002prolog}. These systems generate traceable logical representations, grounding a final stage of logically sound reasoning, giving benefits of interpretability and explainability.

To overcome these limitations, we develop the Formal Semantic Logic Inferer (FSLI) -- a symbolic system that combines semantic parsing with a dedicated logic inference module -- with semantic rules and axioms to specifically solve one-dimensional ordering problems. Unlike hybrid methods, FSLI focuses on precise semantic representation and logical consistency without relying on LLM-generated outputs, through a series of preprocessing, semantic parsing, and logical inference components (Sections \ref{chap:scope-preprocessing}-\ref{chap:logic}).

We evaluate FSLI on  the BIG-bench logical\_deduction benchmark \cite{srivastava2023beyond}, where each problem has a unique valid sequence. FSLI achieves perfect accuracy, confirming the soundness of its semantic parsing and inference pipeline (Table~\ref{tab:performance}).
To assess generalization beyond synthetic tasks, we also evaluate on the much more difficult LSAT-simp-true dataset -- a manually simplified version of LSAT-orig-true that preserves linguistic and logical richness while reducing syntactic variability. On this benchmark, FSLI achieves 88.37\% accuracy, substantially outperforming current approaches. 

Together, these evaluations show  the system’s performance across synthetic and real-world-inspired logical deduction problems, highlighting the potential of logic linguistic computational systems to perform sound and high accuracy natural language logical deduction.  This approach suggests logical computational linguistics approaches for natural language ordering applications such as scheduling, where logically consistent and sound solutions are necessary for planning.

\section{Current Approaches}
\label{sec:relevantlit}

Logical deduction is a complex task requiring structured reasoning such as logical inference. While LLMs excel at many natural language tasks, they rely on pattern-based reasoning rather than explicit inference \cite{marcus2019rebooting, bender-koller-2020-climbing}, limiting their adaptability to novel reasoning tasks \cite{lake2017building, zhang2022paradox}. 

Surveys by \citet{liu2024puzzlesurvey} and \citet{liu2025logicalsurvey} confirm LLMs’ limited robustness and generalization in structured reasoning tasks. LLMs also struggle with formal logic. \citet{parmar2024logicbench, xu2025logicalreasoning} report consistent failures across logic types, especially with negation and multi-step inference. \citet{tyen-etal-2024-llms} find that LLMs are unable to detect their own reasoning errors.

Recent work also demonstrates that LLMs are brittle under perturbation. \citet{hoppe2025robustness, jiang-etal-2024-peek} find that small semantic changes degrade performance, while \citet{mirzadeh2025gsmsymbolic} observe similar failures in mathematical reasoning.

Prompting strategies like Chain of Thought (CoT) \cite{Suzgun2022ChallengingBT} offer partial improvements but vary across tasks. Hybrid systems such as Logic-LM \cite{PanLogicLM23} and Logic-LM++ \cite{kirtania-etal-2024-logic} use external solvers but depend on LLMs to produce reliable logical forms. 

Since structured representations are needed for deductive reasoning, it is crucial to perform the step of semantic parsing, mapping sentences to logical forms.
Combinatory Categorial Grammar (CCG) \cite{steedman2000syntactic} maps syntax to logic \cite{Bos2004WideCoverageSR, Abzianidze2017TowardsUS}, though its large category inventory \cite{hockenmaier2007ccgbank} and flexible combination rules can introduce ambiguity \cite{Moortgat1997-MOOCTL}. 

Still, machine learning-based semantic parsing under CCG or similar frameworks has enjoyed significant (pre-LLM) research attention (e.g.\ \citealp{Abzianidze2017TowardsUS, lewis2014ccg, misra2016neural, liang2013learning, krishnamurthy2012weakly, cai2013large, artzi-etal-2015-broad, Bos2004WideCoverageSR, Xu2016LSTMSC, stanojevic-etal-2021-modeling}).  Most prior work has focused on the accuracy of resulting sentence-level semantic representations, with computational use limited to structured database retrieval question-answering tasks.  

In our work, we seek to tackle deductive reasoning, which also requires considerable interpretation of the passages (what entities must be reasoned about, and what are their properties?) and integration with appropriate computational logic inference engines (what solutions are logically consistent with the premises?). For our semantic parsing approach, instead of using CCG, we separate syntax and semantics to yield modular, interpretable representations better suited for deduction.

\section{Our Approach: Formal Semantic Logic Inferer (FSLI)}  
\label{sec:fsli}

Our goal is to develop a single, integrated system that combines semantic parsing with logical deduction to solve one-dimensional ordering problems expressed in natural language. This design offers three key benefits: (1) the ability to answer one-dimensional ordering queries -- such as `what is second?' -- related to the premises \textbf{without access to the answer choices}; (2) high \textbf{accuracy}; and (3) \textbf{interpretability} (ability to understand each step).

\begin{figure}[!htb] 
    \centering
    \includegraphics[width=\columnwidth]{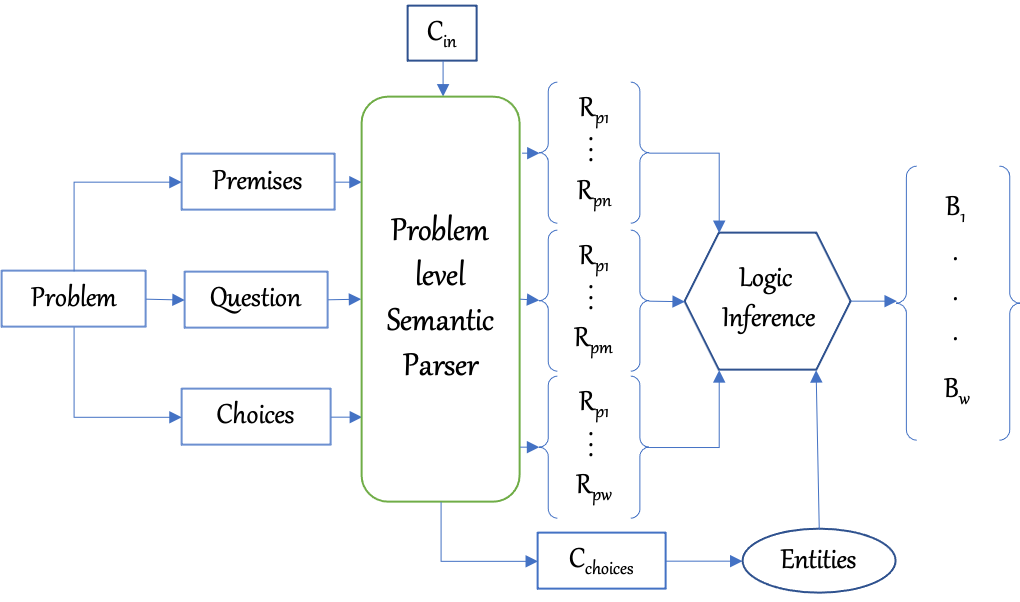}
    \caption{System Overview: FSLI consists of a semantic parser and a logic inference module. $C_{in}$ is the initial context, and $C_{choices}$ contains contexts built from premises, questions, and choices. $R_i$ is the sentence denotation, and $B_i \in \{0, 1\}$ indicates whether the $i$-th choice is a valid deduction.}
    \label{fig:figure-FSLI}
\end{figure}

The computational linguistic and logic system we built -- Formal Semantic Logic Inferer (FSLI) -- embodies two core modules: (1) a formal semantic parsing algorithm (detailed in Section \ref{chap:semantics}) and (2) a first order logic inference engine implemented using logic programming (detailed in Section \ref{chap:logic}). FSLI's system overview is illustrated in Figure \ref{fig:figure-FSLI}. 

We model the semantics of a sentence using compositional formal semantic rules introduced by \citet{Heim1998SemanticsIG} built upon \citet{partee1975toward}'s contributions to \citet{montague1970} Grammar. The result of applying formal semantic rules using word denotations is a logical formula for the sentence. In compositional formal semantics, a context $\mathcal{C}$ is employed to interpret context-dependent expressions such as who the speaker is `I', the place `here',  and the time `now'. In our system, the context $\mathcal{C}$ is represented as mapping between the entity label and the entity description and is dynamically populated.

We algorithmize the compositional formal semantic rules, producing a procedure that converts a sentence into its logical representation using lambda expressions.  Furthermore, we develop an approach to semantic parsing that populates the context with the entities in a dynamic manner. This is needed for logic deduction tasks as the entities are at the core of the problems and need to be extracted as each sentence is parsed. 

Our formal semantic parsing algorithm operates at four levels: word, sentence, paragraph, and problem. The word denotation extraction algorithm uses templates based on its part of speech and named entity extraction to deduce the word's logical representation (denotation). 

The sentence-level semantic parser applies the formal semantic rules, uses the sentence's syntax tree representation to combine the word denotations to the sentence-level denotation, and dynamically extracts the entities in the process. 

The paragraph-level parser keeps track of the entities introduced in each sentence and passes them along to the following sentence in the paragraph. The problem-level semantic parser passes the entities introduced in the premises to the statements when arriving at their semantic denotation.

\section{Datasets and Text Preprocessing}
\label{chap:scope-preprocessing}

We use three datasets to evaluate the task of logical deduction in one-dimensional ordering tasks: the synthetic BIG-bench logical\_deduction dataset (BB), an extracted LSAT dataset \textit{LSAT-orig-true}, and \textit{LSAT-simp-true} a syntactically simplified version of LSAT-orig-true. In all cases, the task involves arranging entities in one-dimension based on a set of constraints such as: \textit{X before Y}, \textit{X immediately before Y}, and \textit{X is first}.

 We evaluate FSLI on \textit{BB} and \textit{LSAT-simp-true} to focus on core logical deduction (Section \ref{sec:big_bench} and \ref{sec:lsat}). We test LLMs on all three datasets to compare performance across different levels of linguistic complexity and logical deduction.

\subsection{BIG-Bench Logical Deduction Dataset}
\label{sec:big_bench}

\citet{srivastava2023beyond}'s \textit{BIG-Bench} is a comprehensive benchmark designed to test the performance and limitations of large language models. Among its many tasks is a logical-deduction benchmark that employs linear arrangement as the core relationship among entities. 

The \textit{logical\_deduction} benchmark consists of three tasks of increasing relational tracking difficulty: {BB\_3}, with three entities; {BB\_5}, with five entities; and {BB\_7}, with seven entities.

Each task consists of a premise, a question, and a set of candidate statements. The candidate statements can either be true or false, with only one correct answer. A problem from BB\_3 is shown in Example \ref{ex:bigbench}.

\subsection{LSAT Logic Games – One-Dimensional Ordering}
\label{sec:lsat}

The \textit{LSAT-orig-true} dataset, extracted from the AR-LSAT corpus \cite{Wang2021FromLT}, consists of one-dimensional ordering problems extracted from the LSAT exam. These problems are both linguistically and logically complex, often allowing multiple valid sequences (e.g., \textit{``A before B before C''} or \textit{``A before C before B''}). Solving them requires advanced deductive reasoning over intricate constraints while navigating syntactically complex language.

The \textit{LSAT-simp-true} dataset is a manually simplified version of \textit{LSAT-orig-true}, created through a \textbf{systematic process} to reduce syntactic complexity while preserving core and logical structures. The resulting problems remain challenging from a semantic and logical deduction standpoint, but minimize confounds related to language comprehension.

Manual simplification applies several operations: \textbf{replacing complex words}, \textbf{splitting long sentences}, \textbf{removing unnecessary phrases}, and \textbf{inserting clarifying words} to indicate constraints or relationships. The resulting \textit{LSAT-simp-true} dataset provides a syntactically simplified setting for evaluating logical deduction while preserving the original logical challenges. 

The problems in both datasets include a {premise}, a {question}, and a set of {answer choices}. The premise introduces the entities and constraints. The question may ask: \textbf{What is true?} (true in at least one valid sequence), \textbf{What must be true?} (true in all valid sequences), or \textbf{What is not true?} (false in all valid sequences). Example \ref{ex:lsat} presents a problem from \textit{LSAT-orig-true} along with its simplified version from \textit{LSAT-simp-true}.

The simplified version in \textit{LSAT-simp-true} preserves the semantic and logical complexities of the original while presenting constraints in a clearer and more direct form.

In Example \ref{ex:lsat}, if the question had been \textit{“What must be true?”}, a different answer set would be needed -- focusing on conclusions that hold across all valid orderings. These questions are often more complex, as they require deducing universally true possibilities. This highlights that the challenge lies in deductive reasoning over constraints, not just understanding the language. \textit{LSAT-simp-true} isolates this reasoning process by minimizing syntactic complexity while preserving logical difficulty.

\begin{tcolorbox}[
  colback=white!10,
  colframe=gray!75!black,
  colbacktitle=white,
  sharp corners=all,
  boxrule=0.1mm,
  left=-0.1mm,
  enhanced,
  breakable
]
\small

{\centering
\refstepcounter{example}\label{ex:lsat}
\textbf{\underline{Example~{\theexample}} \\ }  \vspace{1em}
} 
\textbf{From LSAT-orig-true} \\ \\
\textbf{Premise:} Each of seven television programs—H, J, L, P, Q, S, V—is assigned a different rank: from first through seventh (from most popular to least popular). The ranking is consistent with the following conditions: J and L are each less popular than H. J is more popular than Q. S and V are each less popular than L. P and S are each less popular than Q. S is not seventh.\\

\textbf{Question:} If V is more popular than Q and J is less popular than L, then which one of the following could be true of the ranking?\\

\textbf{Answer Choices:} {[}..., \underline{`J is more popular than V.'}, ...{]}

\tcblower

\small
\textbf{From LSAT-simp-true} \\ \\ 
\textbf{Premise:} There are seven TV programs: H, J, L, P, Q, S, and V.  
J is less popular than H.  
L is less popular than H.  
J is more popular than Q.  
S is less popular than L.  
V is less popular than L.  
P is less popular than Q.  
S is less popular than Q.  
S is not seventh.\\

\textbf{Question:} V is more popular than Q. J is less popular than L. What is true?\\

\textbf{Answer Choices:} {[}..., \underline{`J is more popular than V.'}, ...{]}
\end{tcolorbox}

\subsection{Text Preprocessing}

We use \citet{qi2020stanza}'s Stanford Stanza pipeline for text preprocessing and constituency parsing, including part-of-speech (PoS) tagging, named entity recognition (NER), lemmatization, and morphosyntactic tagging. We apply lemmatization \cite{jurafskyspeech} to extract the base form (lemma) of each word, removing tense and other features irrelevant to logical deduction.

We focus on constituency parsing to support compositional semantics. PoS tags inform denotations for leaves, and the daughters of each node combine to form the denotation of the parent \cite{Heim1998SemanticsIG}. We binarize each parse into Chomsky normal form (CNF) \cite{chomsky1959formal} using NLTK \cite{bird2009natural}, ensuring that every constituent has one or two children. We dynamically manage entity context via description and label assignment (Section~\ref{subsec:sentence_level}).

We map common one dimensional ordering phrases -- such as direction, speed, expense, time, popularity, quantity, and performance -- into logical predicates. These mappings were derived from 10\% of the data and extended using world knowledge (Table~\ref{tab:translations}, Appendix~\ref{appen:translations}).

\section{Semantic Parsing Algorithm}
\label{chap:semantics}

Compositional formal semantics \cite{Heim1998SemanticsIG} is a logical approach to semantics that labels words according to their first-order lambda logic notation. The logic representation of a sentence is determined by combining the logic representations of its constituents using the structure of the syntactic parse.

Logic representation of a word, phrase, or sentence is referred to as the denotation in compositional formal semantics. A denotation consists of two parts: first-order lambda logic notation and the semantic type. \sembrac{Y} represents a denotation of Y where Y could be a word, phrase, or sentence. 

Semantic types indicate whether the sentence, phrase, or word represents an entity, a truth value, or a functional type. An entity is represented with $e$, and a truth value is represented with $t$. $e$ and $t$ are the basic semantic types that can be combined to create functional types. A functional type \tset{$\sigma$,$\tau$} where $\sigma$ and $\tau$ can be a basic type or a functional type.  
An entity is an object such as a noun. Examples of entities include a green apple, a blue banana, and Harry.  

We generate denotations for a problem at three levels (lowest to highest): word level, sentence level, and paragraph level. Lower-level denotations inform the denotations at the higher level. 

\subsection{Word Level Semantic Denotation}
\label{sec:word_level_semantics}

To minimize ambiguity in word denotation, we assign a single denotation per word/word class based on the most probable representation given the domain. For example, in our case, verbs are assumed to be transitive verbs.

We use predicates in our representations to allow the semantic parsing output to be inputted to the logic inference system with little to no modifications. A predicate is a function that can take in one or more inputs; for example, the predicate $before(a, b)$ means \textit{a} is before \textit{b}.  In addition, we extend the denotation types to include two cross-categorial types: \tset{$\mu$,$\mu$} and \tset{$f$,$\mu$}. $\mu$ is a placeholder type that could be of any semantic type, and $f$ is a placeholder type of a functional semantic type. These cross-categorial types provide flexibility by allowing a variety of input types rather than restricting it to a single type for word denotation or introducing ambiguity by allowing more than one possible denotation for a word.

Open-class words with PoS such as nouns and verbs follow general denotation templates, whereas closed-class words -- such as determiners, negation, and auxiliary verbs like `is' and `are' -- have specific, word-level denotations.  Additional denotation templates are provided in Appendix~\ref{sec:denotation_templates}. For instance, the denotation of a noun  follows the template ($\lambda x$: $lemma(x)$, \tset{e,t}). As an example: \sembrac{desk} = ($\lambda x$: $desk(x)$, \tset{e,t}); \sembrac{is} = ($\lambda f$: $f$, \tset{$\mu$,$\mu$}); and  \sembrac{not} = ($\lambda f$: $\_not(f)$, \tset{\tset{e,t}, \tset{e,t}}).

\subsection{Sentence Level Semantic Parser}
\label{subsec:sentence_level}

We address semantic parsing in a rule-based manner, which models  rules introduced in Compositional Formal Semantics \cite{Heim1998SemanticsIG}.  Word denotations are derived using their PoS and NER tags informed using the combination rules. 

The parser maintains a context  $\mathcal{C}$, which  maps entity descriptions to their label. This context is updated dynamically as new entities are encountered during parsing. For example, in the context $\mathcal{C}$  = \{$h$ : `harry(h)', $r$: `ron(r)'\},  `Harry' is assigned the label $h$ and `Ron' is assigned the label $r$. 

In order to identify the entities and use them dynamically across the text, we introduce a special denotation for the definite article (the) and the indefinite articles (a, an) -- detailed in Section \ref{subsub:def_indef}.

This methodology allows the context ($\mathcal{C}$) to be populated as new entities are encountered in each sentence. By populating the context dynamically with the entities, we present a novel approach to extracting entities, particularly for those not captured by traditional NER systems. We assume that no two distinct entities share the same description.

For instance, the final denotation of the sentence `A pink monkey is eating an apple' is $(eat(p, a),\ t)$ with the context  $\mathcal{C}_1 = \{a: apple(a),\ p: pink(p) \land monkey(p)\}$. Detailed derivation of semantically parsing two sentences of a paragraph is provided in  Appendix \ref{appen:sentence_example}.

Let $\mathcal{D}$ be the denotation dictionary, which contains the denotation of the words in the sentence obtained using templates as described in Section \ref{sec:word_level_semantics}. The following formal semantic rules are used to compose the denotations of the words to the denotation of the sentence: 

\begin{enumerate}[left=0pt]
    \item Function Application (FA): If X is a branching node that has two daughters - Y and Z - and if \sembrac{Y} is a function whose domain contains \sembrac{Z}, then \sembrac{$X$} = \sembrac{$Y$}(\sembrac{$Z$})
    \item Non-branching Node (NN): If X is a non-branching node and Y is its sole daughter then \sembrac{$X$} = \sembrac{$Y$} 
    \item Terminal Node (TN): If X is a terminal node (leaf) then \sembrac{$X$} = $\mathcal{D}$[X] if X is not an entity otherwise  \sembrac{$X$} = $\mathcal{C}$[X] if $X$ is an entity
    \item Predicate Modification (PM): If X is a branching node that has two daughters - Y and Z - and if \sembrac{Y} and \sembrac{Z} are of type \tset{e, t}, then \sembrac{X} = ([$\lambda x. $ \sembrac{Y}($x$) and \sembrac{Z}($x$)], \tset{e, t})
\end{enumerate}

\subsection{Definite and Indefinite Article Denotations}
\label{subsub:def_indef}

The indefinite articles `a' and `an' introduce a novelty condition, while the definite article `the' introduces a familiarity condition \cite{abney1987english}. This motivates special predicates: \textbf{\textit{createEntity}} for `a'/`an' and \textbf{\textit{getEntity}} for `the', ensuring entities with the same description share the same label. Let $\mathcal{C}$ be the context:

\begin{itemize}[leftmargin=0cm,rightmargin=0cm] 
\item[] \sembrac{a} = \sembrac{an} = ($\lambda f: createEntity(f, \mathcal{C})$, \tset{\tset{e,t}, e}) 
\item[] \sembrac{the} = ($\lambda f: getEntity(f, \mathcal{C})$, \tset{\tset{e,t}, e})
\end{itemize}

\textbf{\textit{createEntity}} is a predicate that takes a description, creates a label for a new entity, adds it to the context $\mathcal{C}$, and returns the label as its denotation. \textbf{\textit{getEntity}} is a predicate that takes a description and retrieves the corresponding label from $\mathcal{C}$.

\subsection{Paragraph Level Semantic Parser}
\label{subsec:paragraph_level}

A paragraph is the input to the semantic parser, and the output is the denotation of each of the sentences and the context. A context $\mathcal{C}$ is a dictionary that contains the label and description of each entity. 

Let $\mathcal{C}_{in}$ = \{\} be the initial (empty) context. Let the paragraph $P$ = [$X_1$, $X_2$  \dots, $X_n$] where $n$ is the number of sentences in the paragraph. Each sentence $X_i$ is interpreted with respect to the context $\mathcal{C}_{i-1}$ (the output context of $X_{i-1}$), producing a denotation \sembrac{$X_i$} = $R_i$ and an updated context $\mathcal{C}_i$. The denotation of a paragraph is: \sembrac{P} = $[R_1, R_2, \dots, R_n]$ with output context: $\mathcal{C}_{\text{out}} = \mathcal{C}_n$.  An example is illustrated below.

\begin{tcolorbox}[colback=white!10, colframe=gray!75!black, 
colbacktitle=white, sharp corners=all, boxrule=0.1mm, left=-0.1mm]

\small 

Let the two-sentence paragraph $P = [X_1, X_2]$, where:  $X_1$ = `A pink monkey is eating an apple.' and  $X_2$ = `The apple is tasty.' \\ 

Then, the denotation of $P$ is:
\sembrac{P} = [\sembrac{X\_1},\sembrac{X\_2}] = [R\_1, R\_2] = [$(eat(p, a),\ t)$,\ $ (tasty(a),\ t)$]\\ 

The final context is: $\mathcal{C}_{\text{out}} = \mathcal{C}_2 = \{a: apple(a),\ p: pink(p) \land monkey(p)\}$
\end{tcolorbox}

\subsection{Problem Level Semantic Parser}
\label{subsec:problem_level}

A problem contains the premise, a question, and answer choices.
Let $\mathcal{C}_{in}$ = \{\} be the input context. Let $\mathcal{L}$ = ($P$, $Q$, $A$) be the problem where $P$ is the premise that introduces the set of one-dimensional ordering rules, $Q$ is the related question, and $A$ is the candidate statements related to the premise. Let $n$ be the number of sentences in the premise, $m$ the number of sentences in the question, and $w$ the number of sentences of the candidate statements. The denotations of the premise, question, and candidate statements are computed using the paragraph-level semantic parser detailed in Section \ref{subsec:paragraph_level}: 

\begin{itemize}[leftmargin=0cm,rightmargin=0cm]
    \item[] \sembrac{$P$} = [$R_{p1}$, ..., $R_{pn}$] with output context $\mathcal{C}_{premise}$ and input context  $\mathcal{C}_{in}$. 
    \item[] \sembrac{$Q$} = $R_{q}$ with output context $\mathcal{C}_{question}$ and input context $\mathcal{C}_{premise}$. 
    \item[] \sembrac{$A$} = [$R_{a1}$, ..., $R_{aw}$] with output context $\mathcal{C}_{choices}$ and input context $\mathcal{C}_{question}$. 
    \item[] \sembrac{$\mathcal{L}$}  = (\sembrac{$P$}, \sembrac{$Q$}, \sembrac{$A$}), Entities = labels of the entities in $\mathcal{C}_{choices}$. 
\end{itemize}

\begin{table}[htb]
\caption{\label{table:denotation_P_BB} \sembrac{$P_{BB}$} with output context $\mathcal{C}_{P_{BB}}$ =  \{`c': `crow(c)', `q': `quail(q)', `r': `raven(r)'\}}

\resizebox{\columnwidth}{!}{
\begin{tabular}{|l|l|}
\hline
\rowcolor[HTML]{FCE5CD} \textbf{Sentence} & \textbf{Denotation} \\ \hline
A raven, a quail, and a crow & $([r, q, c], e)$ \\ \hline
The quail is the leftmost. & $(position(q, first, 1), t)$ \\ \hline
The raven is the rightmost. & $(last(r), t)$ \\ \hline
\end{tabular}
}
\end{table}

We illustrate the problem-level semantic parser on a BIG-bench logic\_deduction problem (Example \ref{ex:bigbench}). The problem contains the premises $P_{BB}$ = \textbf{premises}, and candidate statements  $A_{{1P}_{BB}}$ = \textbf{candidate statements} with the underlying question $Q_{{1P}_{BB}}$  = `What is true?': 

The problem related to $P_{BB}$ is: $\mathcal{L}_{P_{BB}}$ = ($P_{BB}$, $Q_{{1P}_{BB}}$, $A_{{1P}_{BB}}$). \sembrac{$P_{BB}$} is in Table \ref{table:denotation_P_BB}. Note that the first sentence before the colon is not parsed as it does not provide any needed information for the logic inference.  

The denotation \sembrac{$\mathcal{L}_{P_{BB}}$} = [\sembrac{$P_{BB}$}, [(true(Z), \tset{e,t})], [(position(r, first, 2), t)	,(position(q, first, 2), t)	, (position(c, first, 2), t)]] with the output context $\mathcal{C}_{P_{BB}}$ =  \{`c': `crow(c)', `q': `quail(q)', `r': `raven(r)'\}

\section{Logic Inference}
\label{chap:logic}

We use SWI-Prolog \cite{wielemaker2012swiprolog}, a declarative logic programming language, to encode constraints and infer solutions. The main predicate, $orderInf$, takes a list of entities and constraints and returns a model $\mathcal{M}$ that evaluates logical queries. A model $\mathcal{M}$ is:

\begin{tcolorbox}[
    colback=white!10,
    colframe=gray!75!black,
    sharp corners=all,
    boxrule=0.1mm,
    left=0mm,
    right=0mm,
    top=0mm,
    bottom=0mm,
    enhanced,
    breakable
]
\small

\begin{fleqn}
\begin{equation}
\begin{aligned}
\label{eq:model}
  \mathcal{M} =  &model(Query, Pairs) :   \\
    &orderInf(Entities, Constraints, \\ 
    &\phantom{orderInf }Query, Pairs).
\end{aligned}
\end{equation}
\end{fleqn}

\textbf{\textit{Entities}}: list of entities; \textbf{\textit{Constraints}}: premise-derived rules; \textbf{\textit{Query}}: logical form of question; \textbf{\textit{Pairs}}: resulting ordering.\\

\end{tcolorbox}

The predicate constrains entities to absolute or relative positions in a list, forming a model $\mathcal{M}$ that supports logical queries. Each candidate statement’s semantic form is passed as a query to determine its truth value. Constraints are extracted from the semantic parse, and entities are labeled using the output context $\mathcal{C}_\mathbf{L}$ of problem $\mathcal{L}$.

We formalize one-dimensional ordering using SWI-Prolog’s CLP(FD) library \cite{Triska2012FD}, representing positions as integer indices and enforcing standard axioms like transitivity. Ordering predicates include: \textit{before(A, B)}, \textit{succeed(A, B)}, \textit{not\_(...)}, \textit{first(A)}, \textit{last(A)}, and position-based forms such as \textit{position(A, \textbf{first}, n)}.

To solve $\mathcal{L}_{P_{BB}}$ (Table~\ref{table:denotation_P_BB}, Section~\ref{subsec:problem_level}), the system builds $\mathcal{M}_{BB}$ using Equation~(1) and the semantic parse. The output, [0, 0, 1], indicates that only the third candidate is true.

\begin{tcolorbox}[
    colback=white!10,
    colframe=gray!75!black,
    sharp corners=all,
    boxrule=0.1mm,
    left=0mm,
    right=0mm,
    top=0mm,
    bottom=0mm,
    enhanced,
    breakable
]
{\small
\begin{fleqn}
\begin{equation*}
\begin{aligned}
\label{eq:model}
  \mathcal{M}_{BB}= &model(Query, Pairs) : \\
        &orderInf([r, q, c], (position(q, first, 1), last(r)), \\
        &\phantom{orderInf }Query, Pairs).
\end{aligned}
\end{equation*}
\end{fleqn}
}
\end{tcolorbox}

\section{Performance}
\label{chap:performance} 

\begin{table*}[!htb]
\centering
\setlength{\tabcolsep}{10pt}
\caption{Accuracy (\%) of LLAMA3, GPT4o, o1-preview, and FSLI across datasets by prompt technique, and percentage of no-answer (NR\%) responses from LLAMA3, GPT4o, and o1-preview. (Accuracy, NR\%) }
\label{tab:performance}

\resizebox{\textwidth}{!}{%
\begin{tabular}{|cc|c|c|c|c|c|c|}
\hline
\rowcolor[HTML]{FCE5CD} 
\multicolumn{1}{|c|}{\cellcolor[HTML]{FCE5CD}Model} & Prompt         & BB3                                                           & BB5                                                           & BB7                                                           & BB                                                            & LSAT-simp-true                                                                   & LSAT-orig-true                                                                   \\ \hline
\rowcolor[HTML]{FCE5CD} 
\multicolumn{2}{|c|}{\cellcolor[HTML]{FCE5CD}Size}                   & 300                                                           & 500                                                           & 700                                                           & 1500                                                          & 86                                                                               & 86                                                                               \\ \hline
\multicolumn{1}{|c|}{FSLI}                          & Not Applicable & \textbf{\begin{tabular}[c]{@{}c@{}}(100, --)\end{tabular}} & \textbf{\begin{tabular}[c]{@{}c@{}}(100, --)\end{tabular}} & \textbf{\begin{tabular}[c]{@{}c@{}}(100, --)\end{tabular}} & \textbf{\begin{tabular}[c]{@{}c@{}}(100, --)\end{tabular}} & \textbf{\begin{tabular}[c]{@{}c@{}}(88.37, --)\end{tabular}}                  & \begin{tabular}[c]{@{}c@{}}(--, --)\end{tabular}                              \\ \hline
\multicolumn{1}{|c|}{o1-preview}                    & zero-shot      & \begin{tabular}[c]{@{}c@{}}(99.66, 0.33)\end{tabular}      & \begin{tabular}[c]{@{}c@{}}(99.2, 0.40)\end{tabular}       & \begin{tabular}[c]{@{}c@{}}(98.71, 0)\end{tabular}         & \begin{tabular}[c]{@{}c@{}}(99.06, --)\end{tabular}        & \begin{tabular}[c]{@{}c@{}}(80.23, 1.16)\end{tabular}                         & \begin{tabular}[c]{@{}c@{}}(91.86, 0)\end{tabular}                            \\ \hline
\multicolumn{1}{|c|}{}                              & zero-shot      & \begin{tabular}[c]{@{}c@{}}(98.00, 0)\end{tabular}         & \begin{tabular}[c]{@{}c@{}}(87.60, 0.20)\end{tabular}      & \begin{tabular}[c]{@{}c@{}}(81.28,  0)\end{tabular}         & \begin{tabular}[c]{@{}c@{}}(86.73, --)\end{tabular}        & \begin{tabular}[c]{@{}c@{}}(38.37, 1.16)\end{tabular}                         & \begin{tabular}[c]{@{}c@{}}(29.07,  0)\end{tabular}                            \\ \cline{2-8} 
\multicolumn{1}{|c|}{}                              & one-shot       & \begin{tabular}[c]{@{}c@{}}(92.66, 0)\end{tabular}         & \begin{tabular}[c]{@{}c@{}}(78.6,  0)\end{tabular}          & \begin{tabular}[c]{@{}c@{}}(70.71,  0)\end{tabular}         & \begin{tabular}[c]{@{}c@{}}(77.73, --)\end{tabular}       & \begin{tabular}[c]{@{}c@{}}(27.90, 3.48)\end{tabular}                         & \begin{tabular}[c]{@{}c@{}}(33.72, 1.16)\end{tabular}                         \\ \cline{2-8} 
\multicolumn{1}{|c|}{\multirow{-3}{*}{GPT4o}}       & zero-shot-COT  & \begin{tabular}[c]{@{}c@{}}(100, 0)\end{tabular}           & \begin{tabular}[c]{@{}c@{}}(93.40, 0)\end{tabular}         & \begin{tabular}[c]{@{}c@{}}(82.00, 0.14)\end{tabular}      & \begin{tabular}[c]{@{}c@{}}(89.40,  --)\end{tabular}        & \begin{tabular}[c]{@{}c@{}}(30.53,  1.16)\end{tabular}                         & \begin{tabular}[c]{@{}c@{}}(32.50,  0)\end{tabular}                            \\ \hline
\multicolumn{1}{|c|}{}                              & zero-shot      & \begin{tabular}[c]{@{}c@{}}(96.00, 0.33)\end{tabular}      & \begin{tabular}[c]{@{}c@{}}(91.00, 2.00)\end{tabular}       & \begin{tabular}[c]{@{}c@{}}(76.14, 4.71)\end{tabular}      & \begin{tabular}[c]{@{}c@{}}(85.80, --)\end{tabular}        & \begin{tabular}[c]{@{}c@{}}(36.05, 5.81)\end{tabular}                        & {\color[HTML]{680100} \begin{tabular}[c]{@{}c@{}}(32.56, 13.95)\end{tabular}} \\ \cline{2-8} 
\multicolumn{1}{|c|}{}                              & one-shot       & \begin{tabular}[c]{@{}c@{}}(99.00, 0)\end{tabular}         & \begin{tabular}[c]{@{}c@{}}(92.20, 1.20)\end{tabular}      & \begin{tabular}[c]{@{}c@{}}(76.40, 6.57)\end{tabular}      & \begin{tabular}[c]{@{}c@{}}(86.20, --)\end{tabular}        & {\color[HTML]{680100} \begin{tabular}[c]{@{}c@{}}(34.88, 32.56)\end{tabular}} & {\color[HTML]{680100} \begin{tabular}[c]{@{}c@{}}(43.02, 26.74)\end{tabular}} \\ \cline{2-8} 
\multicolumn{1}{|c|}{\multirow{-3}{*}{LLAMA3}}      & zero-shot-COT  & \begin{tabular}[c]{@{}c@{}}(99.00, 0)\end{tabular}         & \begin{tabular}[c]{@{}c@{}}(93.80, 3.00)\end{tabular}      & {\color[HTML]{680100} \begin{tabular}[c]{@{}c@{}}(59.14, 33.00)\end{tabular}}    & \begin{tabular}[c]{@{}c@{}}(78.67, --)\end{tabular}        & {\color[HTML]{680100} 
 \begin{tabular}[c]{@{}c@{}}(22.09, 34.88)\end{tabular} }                      & {\color[HTML]{680100} \begin{tabular}[c]{@{}c@{}}(30.23, 44.18)\end{tabular}} \\ \hline
\end{tabular}%
}
\end{table*}

We present the performance of the FSLI model and other models on BIG-bench and LSAT data subsets. 

\subsection{Large Language Models}
\label{sec:performance_LLMs}

Unlike FSLI, which requires no data or guidance, LLM evaluations depend on two key factors: (1) separating seen from unseen data, and (2) extracting answers reliably. Because LLMs train on large public corpora (e.g., BIG-bench, LSAT), test overlap is likely. Moreover, structured extraction is essential for consistent outputs.

Creating overlap-free BIG-bench or LSAT variants is impractical. Instead, we assess generalization by manually simplifying questions (Section~\ref{sec:lsat}). Though imperfect, this is more feasible than full dataset creation, which typically requires a team, as with BIG-bench and LSAT. For extraction, we apply a three-stage process: 
\begin{enumerate}[left=0pt] 
\item \textbf{Regex Extraction}: Capture answers from phrases like \texttt{"Correct answer is 2"}. 
\item \textbf{Follow-up Prompt}: Used when regex fails or yields ambiguity. 
\item \textbf{Manual Resolution}: Resolve discrepancies; assign "None" if no answer is given. 
\end{enumerate}

This approach ensures accurate extraction and fair evaluation. Output examples are in Appendix~\ref{appendix:responses}.

\subsection{Results}

\begin{figure}[!htb] 
    \centering
    \includegraphics[width=\columnwidth]{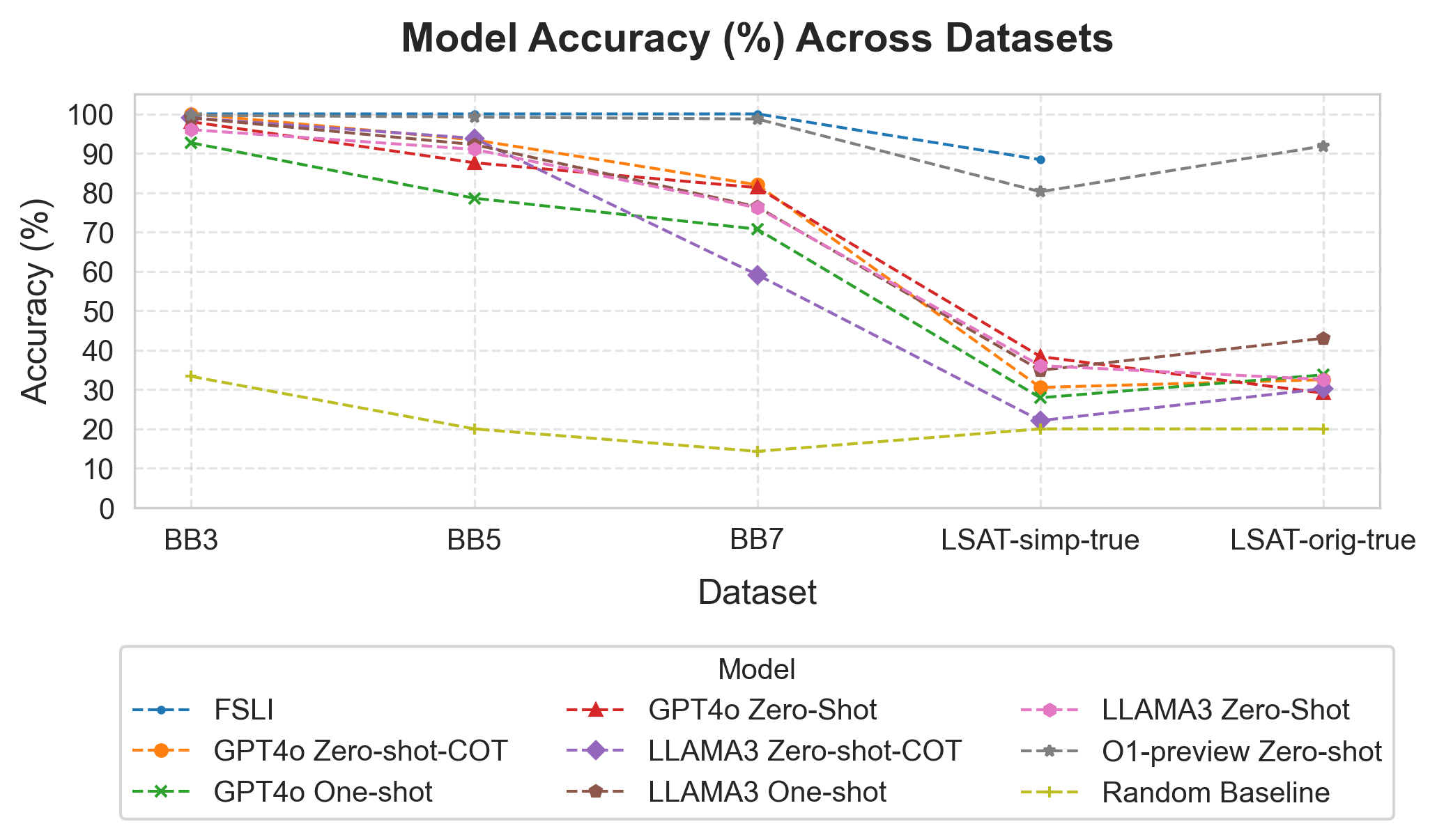}
    \caption{Accuracy trends: FSLI and LLM (o1-preview, GPT4o, and LLAMA3) compared to a random baseline performance. The datasets are ordered based on increasing complexity from left to right}
    \label{fig:performance_anal}
\end{figure}

Table~\ref{tab:performance} presents percent accuracy and no-response rates for FSLI and several LLMs under different prompting strategies. Figure~\ref{fig:performance_anal} shows how accuracy varies with task difficulty (left to right).

We evaluate LLAMA3 and GPT-4o using three prompting methods\footnote{\url{https://www.promptingguide.ai/}}: zero-shot (no guidance), one-shot (a single example with solution) \cite{brown2020language}, and zero-shot Chain-of-Thought (CoT), which encourages step-by-step reasoning \cite{wei2023chainofthought}. Prompt templates are in Appendix~\ref{appendix:prompts}. 
o1-preview is tested only with zero-shot prompting, as it performs internal reasoning via `reasoning tokens’ and is costly to query. It was chosen specifically because its reasoning tokens make it one of the best available models for structured logical tasks -- serving as a strong upper bound for comparison \cite{openai2024o1preview}.
 
\citet{kirtania-etal-2024-logic}’s Logic-LM++ which builts upon  \citet{PanLogicLM23}‘s Logic-LM report 46.32\% accuracy on AR-LSAT, from which LSAT-orig-true is extracted.

\subsection{Discussion of Results}

As the number of answer choices in BIG-bench increases, FSLI maintains 100\% accuracy, while LLM performance deteriorates significantly. FSLI also substantially outperforms all LLMs on LSAT-simp-true and BIG-bench problems. These trends are illustrated in Figure \ref{fig:performance_anal}.

Comparing the last two columns of Table~\ref{tab:performance} (LSAT-orig-true, LSAT-simp-true) shows performance drops in o1-preview (zero-shot), GPT-4o (one-shot, zero-shot CoT), and LLAMA3 (one-shot), but gains in GPT-4o (zero-shot) and LLAMA3 (zero-shot). Overall, simplification tends to reduce LLM performance suggesting limited LLM generalization. 

In addition to lower accuracy, LLAMA3 frequently fails to produces answers, with no-response rates reaching up to 33\% on BB7, 35\% on LSAT-simp-true and 44\% on LSAT-orig-true. In contrast, FSLI always produces a complete output when parsing succeeds, since it operates deterministically without relying on prompt interpretation or generation.

While one-shot generally improves over zero-shot, and zero-shot CoT often outperforms one-shot, exceptions are common. These inconsistencies likely stem from variability in how helpful the prompts are across examples.

\section{Conclusion}

We developed the Formal Semantic Logic Inferer (FSLI), a system grounded in formal semantic parsing and logical inference for solving one-dimensional ordering problems. FSLI achieves 100\% accuracy on BIG-bench’s \textit{logical\_deduction} task and 88.37\% on LSAT-simp-true, outperforming several recent high-performing LLMs including o1-preview, GPT-4o, and LLAMA3.

FSLI has potential applications in domains where task sequencing is critical, such as logistics, project management, and healthcare -- where it could support scheduling, delivery planning, and diagnostic workflows under clinical constraints.

A key direction for future work is extending FSLI to handle one-dimensional ordering with grouping, introducing a second dimension for categorization. This would enable the system to solve more complex scenarios such as event scheduling, timetabling, and production workflows.

Further development may explore temporal reasoning to enable FSLI to interpret time-based expressions and constraints in planning tasks.

These capabilities position FSLI as a foundation for scalable symbolic reasoning systems capable of addressing structured logical deduction tasks across a range of real-world applications.

\section{Limitations}

FSLI is a rule-based system which has been intentionally designed to be modular, interpretable, and achieve formally sound results in both semantic parsing and logical deduction. Our design recognizes the limitations of scope and simplification. Due to FSLI's rule-based nature, we have carefully selected the scope of the system to be one-dimensional ordering problems, which share structural similarities with tasks such as event scheduling, workflows, and other real-world processes involving sequential constraints.

The rule-based nature of FSLI creates the need for non-semantic simplification when handling syntactically complex problems, such as those found in standardized exams like the LSAT. This motivated the creation of LSAT-simp-true, a manually simplified version of LSAT-orig-true, which was extracted from the AR-LSAT dataset. The simplification was necessary to enable effective parsing without altering the underlying logical content. Full details of the datasets and simplification are provided in Section \ref{sec:lsat}.

It is important to note that not all datasets presented require simplification; for example, BIG-bench logical\_deduction is used without preprocessing. The current simplification process is performed manually, but automating it is a central focus of our ongoing work to extend FSLI’s applicability while maintaining its interpretability and formal rigor.

These limitations reflect principled design decisions that prioritize clarity, formal soundness, and trust. FSLI is not intended as a general-purpose solution for question answering, but as a focused and extensible system for logical deduction—particularly valuable in natural language domains where interpretability and logical validity are essential.

\subsection{Integration with Large Language Models}
FSLI currently operates independently of neural pipelines and large language models. However, its deterministic behavior and formal guarantees make it well-suited for integration in hybrid systems -- as a symbolic verifier, constraint enforcer, or correctness filter for LLM-generated outputs.

\section*{Acknowledgements}
We would like to acknowledge the contributions made by Professor Neil Immerman and Professor Katrin Erk. Both of whom have been instrumental in enriching the research. 


\bibliography{main}
\newpage

\appendix
\onecolumn

\section{Denotation Templates}
\label{sec:denotation_templates}

Expanding on Section~\ref{sec:word_level_semantics}, this appendix presents the denotation templates used for both open class and closed class words.

\subsection{Open Class Words}

Focusing on the part of speech categories: noun, adjective, verb, and propernoun; we use templates for the denotations. We use the lemma of a word as part of the logic representation.  The templates are: 
\begin{enumerate}
    \item  \textbf{noun / adjective}  : ($\lambda x$: $lemma(x)$, \tset{e,t}) \\
    which means that a noun is a function of type \tset{e,t} which takes as input an entity and returns a truth value. 
    \item  \textbf{verb} : ($\lambda x$: $\lambda y$: $lemma(y, x)$, \tset{e,\tset{e,t}}) \\ 
    which means that a verb is a function of type \tset{e,\tset{e,t}} which takes as input an entity and returns a function which takes as input an entity and returns a truth value. 
    \item \textbf{propernoun}  : ($\lambda x$: $lemma(x)$, \tset{\tset{e,t}, ec}) \\ 
    which means that a propernoun is a function of type \tset{\tset{e,t}, ec}, which takes as input a function of type \tset{e,t} and creates an entity using the input function. 
\end{enumerate}

We have introduced a special denotation type for propernouns \tset{\tset{e,t},ec}. This follows the syntactic representation that a propernoun could be part of a determiner phrase with a hidden determiner \cite{abney1987english}. This representation makes the system robust against errors in NER and allows for handling two-part propernouns like `New York'. \tset{\tset{e,t},ec} means that an entity is created and placed in the context $\mathcal{C}$ if not previously placed in it. 





\section{Sentence Level Parsing Example}
\label{appen:sentence_example} 

We demonstrate how context is populated and denotations are extracted during sentence-level parsing, using the rules outlined in Section~\ref{subsec:sentence_level}. Consider the two-sentence paragraph: ``A pink monkey is eating an apple. The apple is tasty.''

\begin{figure}[!hbt]
    \center
    \includegraphics[width=.5\columnwidth]{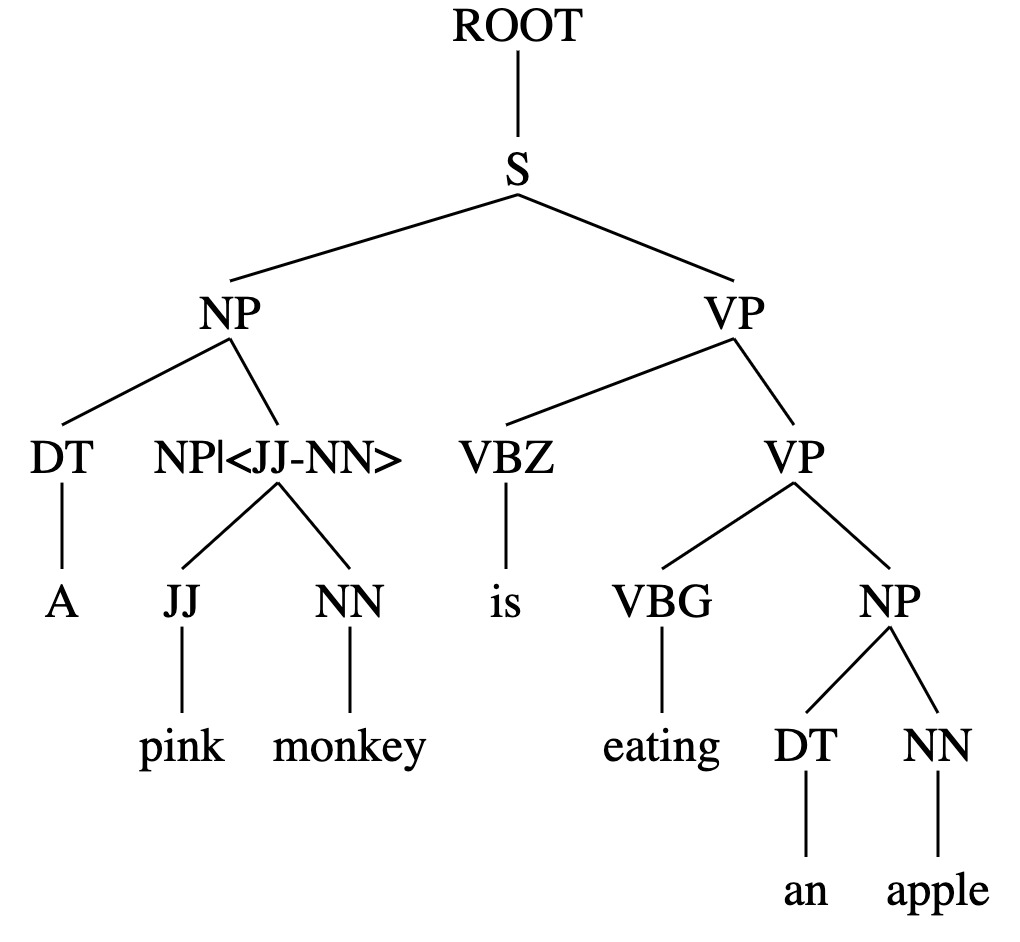}\hfill
    \includegraphics[width=.350\columnwidth]{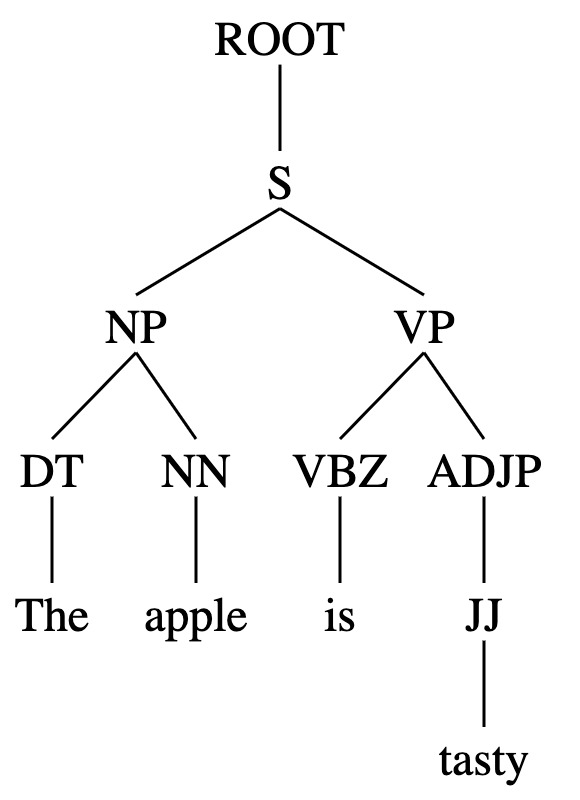}
    \\[\smallskipamount]
    \caption{The binarized syntactic parses of the sentences in `A pink monkey is eating an apple. The apple is tasty.'}
    \label{fig:illustrative_example}
\end{figure}

Each sentence’s syntax tree is traversed bottom-up from right to left to compute its denotation and update the context. This direction is useful, as it prioritizes processing nouns before their modifiers and related verbs. The syntax trees are shown in Figure~\ref{fig:illustrative_example}.

Let $\mathcal{C}$ be the context which starts empty.
The denotation of the first sentence `A pink monkey is eating an apple' is derived as follows:

\begin{enumerate}

\item $\mathcal{C}$ = \{\}, \\
    \sembrac{apple} = ($\lambda x: apple(x)$, \tset{e,t}) \hfill (by TN) \\ 
    \sembrac{an} = ($\lambda f: createEntity(f,\mathcal{C})$, \tset{\tset{e,t}, e}) \hfill (by TN) \\ 
    \sembrac{an apple} = \sembrac{an}(\sembrac{apple}) = a , \\ $\mathcal{C}$ = \{a: apple(a)\} \hfill (by FA) 
    \item $\mathcal{C}$ = \{a: apple(a)\}, \\
    \sembrac{is} = ($\lambda f: f$, \tset{$\mu$,$\mu$})  \hfill (by TN) \\ 
    \sembrac{eating} = ($\lambda x: \lambda y: eat(y, x)$, \tset{e,\tset{e,t}})  \hfill (by TN) \\ 
    \sembrac{is eating} = \sembrac{is}(\sembrac{eating}) = \\  
    ($\lambda x: \lambda y: sing(y, x)$, \tset{e,\tset{e,t}}) \hfill (by FA and $\beta$-reduction) 
    \item $\mathcal{C}$ = \{a: apple(a)\}, \\
    \sembrac{monkey} = ($\lambda x: monkey(x)$, \tset{e,t}) \hfill (by TN) \\ 
    \sembrac{pink} = ($\lambda x: pink(x)$, \tset{e,t}) \hfill (by TN) \\ 
    \sembrac{pink monkey} = ($\lambda x$: $pink(x)$ and $monkey(x)$, \tset{e,t}) \hfill (by PM) 
    \item $\mathcal{C}$ = \{a: apple(a)\}, \\
    \sembrac{a} = ($\lambda f: createEntity(f,\mathcal{C})$, \tset{\tset{e,t}, e}) \hfill (by TN) \\ 
    \sembrac{a pink monkey} = \sembrac{a}(\sembrac{pink monkey}) = p , \\ $\mathcal{C}$ = \{a: apple(a), p: pink(p) and monkey(p)\} \hfill (by FA, step (c)) 
    \item $\mathcal{C}$ = \{a: apple(a), p: pink(p) and monkey(p)\}, \\
    \sembrac{is eating} = ($\lambda x: \lambda y: eat(y, x)$, \tset{e,\tset{e,t}}) \hfill (by step (b))  \\ 
     \sembrac{an apple} = a \hfill (by step (a)) \\
    \sembrac{is eating an apple} = ($\lambda y: eat(y, a)$, \tset{e,t}) \hfill (by FA and $\beta$-reduction) 
    \item $\mathcal{C}$ = \{a: apple(a), p: pink(p) and monkey(p)\}, \\
    \sembrac{is eating an apple} = ($\lambda y: eat(y, a)$, \tset{e,t}) \hfill (by step 5)  \\ 
     \sembrac{a pink monkey} = p \hfill (by step (d)) \\
    \sembrac{a pink monkey is eating an apple} = \sembrac{is eating an apple}(\sembrac{a pink monkey}) = (($\lambda y: eat(y, a)$, \tset{e,t})(p) =  ($eat(p, a)$, t) \hfill (by FA and $\beta$-reduction) 
    
\end{enumerate}

The denotation of the second sentence `The apple is tasty' is derived as follows. The context $\mathcal{C}$  will be the final context derived from the first sentence $\mathcal{C}$ = \{a: apple(a), p: pink(p) and monkey(p)\}:  

\begin{enumerate}

    \item $\mathcal{C}$ = \{a: apple(a), p: pink(p) and monkey(p)\}, \\
    \sembrac{apple} = ($\lambda x: apple(x)$, \tset{e,t}) \hfill (by TN) \\ 
    \sembrac{the} = ($\lambda f$: $getEntity(f, \mathcal{C})$, \tset{\tset{e,t}, e}) \hfill (by TN) \\ 
    \sembrac{the apple} = \sembrac{the}(\sembrac{apple}) = a  \hfill (by FA)  
    \item $\mathcal{C}$  = \{a: apple(a), p: pink(p) and monkey(p)\}, \\
    \sembrac{is} = ($\lambda f: f$, \tset{$\mu$,$\mu$})  \hfill (by TN) \\ 
    \sembrac{tasty} = ($\lambda x:  tasty(x)$, \tset{e,t})   \hfill (by TN) \\ 
    \sembrac{is tasty} = \sembrac{is}(\sembrac{tasty}) = \\  
    ($\lambda x:  tasty(x)$, \tset{e,t})  \hfill (by FA and $\beta$-reduction) 
    \item $\mathcal{C}$ =  \{a: apple(a), p: pink(p) and monkey(p)\}, \\
    \sembrac{the apple} =  a  \hfill (by step (a)) \\ 
    \sembrac{is tasty} = ($\lambda x:  tasty(x)$, \tset{e,t})  \hfill (by step (b)) \\
     \sembrac{the apple is tasty} = \\ 
     \sembrac{is tasty}(\sembrac{the apple}) = \\ 
     ($\lambda x:  tasty(x)$, \tset{e,t})(a) = (tasty(a), t)  \hfill (by FA and $\beta$-reduction) 
\end{enumerate}

\section{LLM Prompt Templates} 
\label{appendix:prompts}

This appendix provides the prompt templates used for each prompting technique evaluated in our experiments (Section \ref{chap:performance}). In each case, the placeholder \texttt{{PROBLEM}} is substituted with a logical reasoning question drawn from either BIG-bench’s \textit{logical\_deduction} task or one of the derived LSAT datasets: \textit{LSAT-orig-true} and \textit{LSAT-simp-true}.

\subsection{Zero-Shot Prompt}
\begin{quote}
please give me the answer to a multiple choice question by giving the index of the correct answer (starting from 0):

\texttt{\{PROBLEM\}}
\end{quote}

\subsection{One-Shot Prompt}
\begin{quote}
premise:  
In an antique car show, there are three vehicles: a convertible, a tractor, and a hatchback. The hatchback is older than the convertible. The tractor is the newest.

question:  
what is true?

answer choices:  
\texttt{[`The convertible is the oldest.', `The tractor is the oldest.', `The hatchback is the oldest.']}

The answer using index (starting from zero) of the correct answer choice is: 2

\texttt{\{PROBLEM\}}

The answer using index (starting from zero) of the correct answer choice is:
\end{quote}

\subsection{Zero-Shot CoT Prompt}
\begin{quote}
A logical deduction task which requires deducing the order of a sequence of objects. Please give me the answer to a multiple choice question by giving the index of the correct answer (starting from 0):

\texttt{\{PROBLEM\}}

Let's think step by step.
\end{quote}

\section{Examples of Model Responses} \label{appendix:responses}

This appendix provides examples of responses from LLAMA3, GPT4o, and o1-preview using zero-shot and zero-shot Chain-of-Thought (CoT) prompting. Some responses contain an answer, while others do not. Section \ref{sec:performance_LLMs} details how we parse these responses to extract the answer (as an index).

\subsection{LLAMA3 Responses}

\paragraph{Example 1: LLAMA3 response using zero-shot CoT prompting (no answer)}  
\begin{quote}

"To solve this, let's analyze the information given step by step: 1. **The water is before the landscaping.** 2. **The power is before the gas.** 3. **The power is before the satellite.** 4. **The gas is not second.** and **The gas is not third.** This means gas can only be fourth, fifth, or sixth. 5. **The satellite is not second.** and **The satellite is not third.** This means satellite can only be fourth, fifth, or sixth. 6. **The telephone is not second.**, **The telephone is not third.**, and **The telephone is not sixth.** This means telephone can only be first, fourth, or fifth. Given these constraints, let's try to establish some order: - Since the power must be before both the gas and the satellite, and both gas and satellite cannot be second or third, power must be either first or second to allow for this arrangement. - Water must be before landscaping, but there's no restriction that directly links water and power, except through the constraints of positions for gas, satellite, and telephone. Considering the options: - **A. The landscaping is earlier than the telephone.** We can't be certain of this because landscaping's position relative to telephone isn't directly constrained. - **B. The power is earlier than the landscaping.** This isn't necessarily true; power could be after landscaping if water is first. - **C. The telephone is earlier than the gas.** Given the constraints, telephone could indeed be before gas, especially since gas can't be second or third, and telephone has the possibility of being fourth. - **D. The telephone is earlier than the water.** This isn't necessarily true; water could be before telephone. - **E. The water is earlier than the gas.** This statement aligns with the constraints since water can be before landscaping, and gas is restricted to later positions. However, upon reevaluation, considering the strict constraints: - Power must be before gas and satellite. - Water must be before landscaping. - Telephone's restrictions place it likely in the first position given the constraints around it, especially since it can't be second, third, or sixth, and considering the placement of other services. Given this, the statement that **"The water is earlier than the gas"** (E) is indeed supported by the constraints since water can be in an earlier position (e.g., first or second) to accommodate the"

\end{quote}

\paragraph{Example 2: LLAMA3 response using zero-shot CoT prompting (with answer)}  
\begin{quote}

"Let's break down the information given: 1. The loquats are less expensive than the cantaloupes. 2. The cantaloupes are less expensive than the apples. 3. The oranges are the most expensive. 4. The apples are the third-most expensive. From these premises, we can deduce the order of expense from least to most as follows: - Since the loquats are less expensive than the cantaloupes, and the cantaloupes are less expensive than the apples, the order for these three is: loquats < cantaloupes < apples. - Given that the oranges are the most expensive and the apples are the third-most expensive, it confirms that there is one more fruit that fits between apples and oranges in terms of expense, but it's not specified which one. However, since the question is about what is true regarding the options provided, we don't necessarily need to identify the exact middle fruit to answer the question about the second-cheapest. - The peaches have not been directly compared to any other fruit in terms of expense, but since the question is about identifying the second-cheapest and we know the loquats are the cheapest from the given comparisons, we look at the options provided. Given the comparisons, the loquats are the cheapest, making the statement about the loquats being the second-cheapest incorrect. The correct sequence from the cheapest based on the information is loquats, then an unspecified fruit (because the direct comparison between peaches and the others isn't provided), then cantaloupes, then apples, then oranges. However, looking at the answer choices provided and considering the logical deductions we've made: - The loquats are the cheapest, not the second-cheapest. - The cantaloupes are less expensive than the apples, placing them before the apples. - The apples are the third-most expensive, meaning there are two fruits less expensive than them, and since the loquats are the least expensive, the cantaloupes must be the second-cheapest based on the provided comparisons. \textbf{Thus, the correct answer is: 3 (The cantaloupes are the second-cheapest}.) This conclusion is based on the logical sequence deduced from the premises provided and the options available for the question."

\end{quote}

\subsection{GPT4o Responses}

\paragraph{Example 1: GPT4o response using zero-shot  CoT prompting (with an answer)}  
\begin{quote}

"To solve this logical deduction task, let's analyze the given premise and order the fruits by their price from most expensive to least expensive. 1. The premise states that the oranges are the most expensive. 2. The apples are the third-most expensive. From these two clues, we know the order starts with: - 1st: Oranges - 2nd: Unknown - 3rd: Apples - 4th: Unknown - 5th: Unknown Now, let's consider the information about loquats and cantaloupes: 3. The loquats are less expensive than the cantaloupes. 4. The cantaloupes are less expensive than the apples. This information helps us place cantaloupes and loquats in relation to apples: - Since cantaloupes are less expensive than apples and loquats are less expensive than cantaloupes, neither cantaloupes nor loquats can occupy the 2nd position. Therefore, cantaloupes and loquats must occupy the 4th and 5th positions respectively. So far, our sequence is: - 1st: Oranges - 2nd: Unknown - 3rd: Apples - 4th: Cantaloupes - 5th:Loquats This leaves peaches as the only fruit that can be the 2nd-most expensive. Therefore, the full order starting with the most expensive is: 1. Oranges 2. Peaches 3. Apples 4. Cantaloupes 5. Loquats Given this order, the second-cheapest fruit is the 4th-most expensive fruit, which is cantaloupes. So, the answer to the multiple-choice question is: - 'The cantaloupes are the second-cheapest.' \textbf{The index of this answer choice is 3}."

\end{quote}

\paragraph{Example 2: GPT4o response using zero-shot  CoT prompting (without an answer)} \begin{quote}
To solve this, we'll need to deduce the relative order of the fruits from least to most expensive using the given information.

1. \textbf{Identify the known rankings:}  
   - Mangoes are the fourth-most expensive.  
   - Pears are the third-most expensive.  
   - Loquats are the third-cheapest.  

2. \textbf{Use the relational price clues:}  
   - Apples are less expensive than cantaloupes.  
   - Apples are more expensive than peaches.  
   - Peaches are more expensive than plums.  

Based on these clues, let's try to construct the order:  

- \textbf{Start placing known positions:}  
  - Since loquats are the third-cheapest, there are two fruits cheaper than loquats and four more expensive.  
  - Mangoes fall into the fourth-most expensive position, with three fruits less expensive and three more expensive.  
  - Pears are third-most expensive, meaning two fruits are cheaper and four are more expensive.  

- \textbf{Now test possible placements for other fruits:}  
  - If loquats are 3rd cheapest (5th overall from the most expensive side), placing plums as cheaper than loquats fits our requirement.  
  - Peaches must be more expensive than plums but less than apples.  

- \textbf{Deriving the set:}  
  - Fruits cheaper than loquats (2 positions) are plums and one other.  
  - Peaches are more expensive than plums but cheaper than both loquats and apples.  
  - The possible order of increasing cost makes plums the least, then something, then loquats, then apples.  

Now we can begin to rank:  
a. \textbf{From expensive to cheapest:}  
   1. Cantaloupes (Since pears are 3rd and mangoes are 4th, the only logical place for unmentioned cantaloupes is 1st or 2nd, but they need to enable pears to be 3rd when mangoes are 4th)  
   2. Something (either apples or potentially something we previously missed but we need more indication for apples)  
   3. Pears  
   4. Mangoes  
   5. Loquats  
   6. Peaches  
   7. Plums  

Therefore, with all logical deductions, the tentative options:  
- Peaches, plums, and plums after loquats being 3rd cheapest implies either apples or cantaloupes are filling the 2nd.  

To finalize, test cantaloupes over the others but apples based on the inability of cantaloupes to be second and option hints above makes them not sufficiently more known.  

This is allowing us to consider unmentioned ties that infer \textit{"The cantaloupes are the second-cheapest."}  

So, checking the option provided: The correct answer is that the fruit in question must then end with relation A to here:  

- Loquats fill the central part, but apples checkback second the next logical step given loquats non-verification needing cantaloupes tied regardless of reconfirmation, not fitting pure or plus apple contention for the chain. Outcomes like pear control which was 3rd.  

Answer index is therefore based on question assignment choice repeat tie like cantaloupes or peaches implying without fully correcting the outside false flag-created potential abstracts on refill so:  
1. \texttt{0. The cantaloupes are the second-cheapest.}  

Through direct primary position form verification, it implies this cannot be remanaged with the chosen scenario of options above.
\end{quote}

\subsection{o1-preview Responses}

\paragraph{Example 1: o1-preview response using zero-shot prompting (no answer)}  
\begin{quote}
Let's break down the information given:

1. **Pears are less expensive than plums:**  
   - This means **pears** cost less than **plums**.  
   - In terms of price: **Pears < Plums**  

2. **Kiwis are less expensive than pears:**  
   - This means **kiwis** cost less than **pears**.  
   - Updating our price order: **Kiwis < Pears**

Combining both pieces of information, we can establish the overall price order:  
- **Kiwis < Pears < Plums**

So, the question is asking: **What is true?**

Given the order we've determined, which fruit is the cheapest among the three?  
Consider the options and see which one aligns with our findings.
\end{quote}

\paragraph{Example 2: o1-preview response using zero-shot prompting (contains an answer)}  
\begin{quote}
The correct answer is index **3**. Here's the reasoning:

1. **Mya** finished first.  
2. **Dan** finished last.  
3. **Amy** finished above **Mel**.  
4. **Amy** finished below **Eli**.  

This gives us the following order:  

1. **Mya**  
2. **Eli**  
3. **Amy**  
4. **Mel**  
5. **Dan**  

So, **Eli finished second**, \textbf{which corresponds to index 3} in the answer choices.
\end{quote}

\paragraph{Example 3: o1-preview response using zero-shot prompting (contains an answer)}  
\begin{quote}
`1'
\end{quote}

\section{Translations}
\label{appen:translations}

Table \ref{tab:translations} enumerates some of the phrases that are used to indicate one-dimensional ordering.

\begin{table*}[!hbt]
\caption{Selected translations}
\label{tab:translations}
\resizebox{\textwidth}{!}{%
\begin{tabular}{|l|l|l|l|l|}
\hline
\rowcolor[HTML]{D3D3D3} 
\textbf{Category} & \textbf{Possible Preceding Phrase} & \textbf{Keyword} & \textbf{Possible Succeeding Phrase} & \textbf{Translation} \\ \hline
\multirow{4}{*}{Direction} 
    & to the           & right       & of           & after  \\ \cline{2-5} 
    & to the           & left        & of           & before \\ \cline{2-5} 
    &                  & rightmost   &              & last   \\ \cline{2-5} 
    &                  & leftmost    &              & first  \\ \hline
\multirow{4}{*}{Relative Position} 
    & the              & first from the left  &          & first \\ \cline{2-5} 
    & the              & third from the left  &          & third \\ \cline{2-5} 
    & the              & first from the right &          & last  \\ \cline{2-5} 
    & the              & third from the right &          & thirdL \\ \hline
\multirow{5}{*}{Performance} 
    &                  & finished above  &             & before \\ \cline{2-5} 
    &                  & finished below  &             & after  \\ \cline{2-5} 
    &                  & finished third  &             & third  \\ \cline{2-5} 
    &                  & finished first  &             & first  \\ \cline{2-5} 
    &                  & finished last   &             & last   \\ \hline

\end{tabular}%
}
\end{table*}

\twocolumn

\end{document}